\documentclass[10pt, conference, compsocconf]{IEEEtran}
\usepackage{times}
\usepackage{epsfig}
\usepackage{graphicx}
\usepackage{amsmath}
\usepackage{amsthm}
\usepackage{amssymb}

\usepackage{amsfonts}
\usepackage{algorithm}
\usepackage{algorithmic}

\newtheorem{theorem}{Theorem}
\newtheorem{definition}{Definition}

\begin{document}

\title{NESVM: a Fast Gradient Method for Support Vector Machines}

\author{Tianyi Zhou, Dacheng Tao, Xindong Wu}

\maketitle

\begin{abstract}
Support vector machines (SVMs) are invaluable tools for many practical applications in artificial intelligence, e.g., classification and event recognition. However, popular SVM solvers are not sufficiently efficient for applications with a great deal of samples as well as a large number of features. In this paper, thus, we present NESVM, a fast gradient SVM solver that can optimize various SVM models, e.g., classical SVM, linear programming SVM and least square SVM. Compared against SVM-Perf \cite{SVM_Perf}\cite{PerfML} (its convergence rate in solving the dual SVM is upper bounded by $\mathcal O(1/\sqrt{k})$, wherein $k$ is the number of iterations.) and Pegasos \cite{Pegasos} (online SVM that converges at rate $\mathcal O(1/k)$ for the primal SVM), NESVM achieves the optimal convergence rate at $\mathcal O(1/k^{2})$ and a linear time complexity. In particular, NESVM smoothes the non-differentiable hinge loss and $\ell_1$-norm in the primal SVM. Then the optimal gradient method without any line search is adopted to solve the optimization. In each iteration round, the current gradient and historical gradients are combined to determine the descent direction, while the Lipschitz constant determines the step size. Only two matrix-vector multiplications are required in each iteration round. Therefore, NESVM is more efficient than existing SVM solvers. In addition, NESVM is available for both linear and nonlinear kernels. We also propose ``homotopy NESVM'' to accelerate NESVM by dynamically decreasing the smooth parameter and using the continuation method. Our experiments on census income categorization, indoor/outdoor scene classification£¬ event recognition and scene recognition suggest the efficiency and the effectiveness of NESVM. The MATLAB code of NESVM will be available on our website for further assessment.
\end{abstract}


\IEEEpeerreviewmaketitle

\section{Introduction}

Support Vector Machines (SVMs) are prominent machine learning tools for practical artificial intelligence applications \cite{GilbertSVM}\cite{Walk}. However, existing SVM solvers are not sufficiently efficient for practical problems, e.g., scene classification and event recognition, with a large number of training samples as well as a great deal of features. This is because the time cost of working set selection or Hessian matrix computation in conventional SVM solvers rapidly increases with the slightly augmenting of the data size and the feature dimension. In addition, they cannot converge quickly to the global optimum. Recently, efficient SVM solvers have been intensively studied on both dual and primal SVMs.

Decomposition methods, e.g., sequential minimal optimization (SMO) \cite{SMO}, LIBSVM \cite{LibSVM} and SVM-Light \cite{SVM_Light}, were developed to reduce the space cost for optimizing the dual SVM. In each iteration round, they consider a subset of constraints that are relevant to the current support vectors and optimize the corresponding dual problem on the selected working set by casting it into a quadratic programming (QP) problem. However, they are impractical to handle large scale problems, because their time complexities are super linear in $n$ and the maximization of the dual objective function leads to a slow convergence rate to the optimum of the primal objective function.

Structural SVM, e.g., SVM-Perf \cite{SVM_Perf}\cite{PerfML}\cite{TJSVM}, is recently proposed to improve the efficiency of optimization on the dual SVM. It reformulates the classical SVM into a structural form. In each iteration round, it firstly computes the most violated constraint from the training set by using a cutting-plane algorithm and adds this constraint to the current working set, then a QP solver is applied to optimize the corresponding dual problem. The Wolfe Dual of structural SVM is sparse and thus the size of each QP problem is small. It has been proved that the convergence rate of SVM-Perf is upper bounded by $\mathcal O(1/\sqrt{k})$ and a lot of successful applications show the efficiency of SVM-Perf. However, it cannot work well when classes are difficult to be separated, e.g., the overlap between classes is serious or distributions of classes are seriously imbalanced. In this scenario, a large $C$ is required to increase the support vectors and thus it is inefficient to find the most violated constraint in SVM-Perf.

Many recent research results \cite{Primal_SVM} show advantages to solve the primal SVM on large scale datasets. However, it is inefficient to directly solve the corresponding QP of the primal SVM if the number of constraints is around the number of samples, e.g., the interior point method for solving the primal SVM. One available solution is to write each of the constraints to the objective function as a hinge loss $\hbar$ and reformulate the problem as an unconstrained one.

Let $X\in \mathbb R^{n\times p}$ and $y\in \mathbb R^n$ be the training dataset and the corresponding label vector, respectively, where the vector $X_i\in \mathbb R^p$ is the $i^{th}$ sample in $X$ and $y_i\in \left\{1,-1\right\}$ is the corresponding label. Let the weight vector $w$ be the classification hyper-plane. The reformulated primal problem of classical SVM is given by
\begin{align}
&\min_{w\in \mathbb R^p} F(w)=\frac{1}{2}\|w\|_2^2+C\sum_{i=1}^n \hbar \left(y_iX_i,w\right),\\
&\hbar \left(y_iX_i,w\right)=\max\left\{0,1-y_iX_iw\right\}.
\end{align}
Since the hinge loss is non-differentiable, first order methods, e.g., subgradient method and stochastic gradient method, can achieve the solution with the convergence rate $\mathcal O(1/\sqrt{k})$, which is not sufficiently fast for large-scale problems. Second order methods, e.g., Newton method and Quasi-Newton method, can obtain the solution as well by replacing the hinge loss with differentiable approximations, e.g., $\max\left\{0,1-y_iX_iw\right\}^q$ used in \cite{Primal_SVM} or the integral of sigmoid function used in \cite{smooth_SVM}. Although, the second order methods achieve the optimal convergence rate at $\mathcal O(1/k^2)$, it is expensive to calculate the Hessian matrix in each iteration round. Therefore, it is impractical to optimize the primal SVM by using the second order methods.

Recently, Pegasos \cite{Pegasos}, a first order online method, was proposed by introducing a projection step after each stochastic gradient update. It converges at rate $\mathcal O(1/k)$. In addition, its computational cost can be rapidly reduced if the feature is sparse, because the computation of the primal objective gradient can be significantly simplified. Therefore, it has been successfully applied to document classification. However, it hardly outperforms SVM-Perf when the feature is dense, which is a frequently encountered situation in artificial intelligence, e.g., computer vision tasks.

In this paper, we present and analyze a fast gradient SVM framework, i.e., NESVM, which can solve the primal problems of typical SVM models, i.e., classical SVM (C-SVM) \cite{CSVM}, linear programming SVM (LP-SVM) \cite{LPSVM} and least square SVM (LS-SVM) \cite{LSSVM}, with the proved optimal convergence rate $\mathcal O(1/k^{2})$ and a linear time complexity. The ``NES'' in NESVM refers to Nesterov's method to acknowledge the fact that NESVM is based on the method. Recently, Nesterov's method has been successfully applied to various optimization problems \cite{YeAccelerateNuclearNorm}\cite{NesterovLogistic}, e.g., compressive sensing, sparse covariance selection, sparse PCA and matrix completion. The proposed NESVM smoothes the non-differentiable parts, i.e., hinge loss and $\ell_1$-norm in the primal objective functions of SVMs, and then uses a gradient-based method with the proved optimal convergence rate to solve the smoothed optimizations. In each iteration round, two auxiliary optimizations are constructed, a weighted combination of their solutions is assigned as the current SVM solution, which is determined by the current gradient and historical gradients. In each iteration round, only two matrix-vector multiplications are required. Both linear and nonlinear kernels can be easily applied to NESVM. The speed of NESVM remains fast when dealing with dense features.

We apply NESVM to census income categorization \cite{UCI}, indoor scene classification \cite{Indoor_scene}, outdoor scene classification \cite{Outdoor_scene} and event recognition \cite{Events} on publicly available datasets. In these applications, we compare NESVM against four popular SVM solvers, i.e., SVM-Perf, Pegasos, SVM-Light and LIBSVM. Sufficient experimental results indicate that NESVM achieves the shortest CPU time and a comparable performance among all the SVM solvers.


\section{NESVM}

We write typical primal SVMs in the following unified form:
\begin{align}
\min_{w\in \mathbb R^p} F(w)=R(w)+C\cdot L(y_iX_i, w),
\label{equ:SVM}
\end{align}
where $R(w)$ is a regularizer inducing the margin maximization in SVMs, $L(y_iX_i, w)$ is a loss function for minimizing the classification error, and $C$ is the SVM parameter. For example, $R(w)$ is the $\ell_2$-norm of $w$ in C-SVM, $R(w)$ is the $\ell_1$-norm of $w$ in LP-SVM, $L(y_iX_i, w)$ is the hinge loss of the classification error in C-SVM, and $L(y_iX_i, w)$ is the least square loss of the classification error in LS-SVM. It is worth emphasizing that nonlinear SVMs can be unified as Eq.\ref{equ:SVM} as well. Details are given at the end of Section 2.3.

In this section, we introduce and analyze the proposed fast gradient SVM framework, i.e., NESVM, based on Eq.\ref{equ:SVM}. We first show that the non-differentiable parts in SVMs, i.e., the hinge loss and the $\ell_1$-norm, can be written as saddle point functions and smoothed by subtracting respective prox-functions. We then introduce Nesterov's method \cite{Nesterov} to optimize the smoothed SVM objective function $F_\mu(w)$. In each iteration round of NESVM, two simple auxiliary optimizations are constructed and the optimal linear combination of their solutions is adopted as the solution of Eq.\ref{equ:SVM} at the current iteration round. NESVM is a first order method and achieves the optimal convergence rate of $\mathcal O(1/k^2)$. In each iteration round, it requires only two matrix-vector multiplications. We analyze the convergence rate and time complexity of NESVM theoretically. An accelerated NESVM using continuation method, i.e., ``homotopy NESVM'' is introduced at the end of this section. Homotopy NESVM solves a sequence of NESVM with decreasing smooth parameter to obtain an accurate approximation of the hinge loss. The solution of each NESVM is used as the ``warm start'' of the next NESVM in homotopy NESVM and thus the computational time for each NESVM can be significantly saved.

\subsection{Smooth the hinge loss}

\begin{figure}[t]
\begin{center}
 \includegraphics[width=0.85\linewidth]{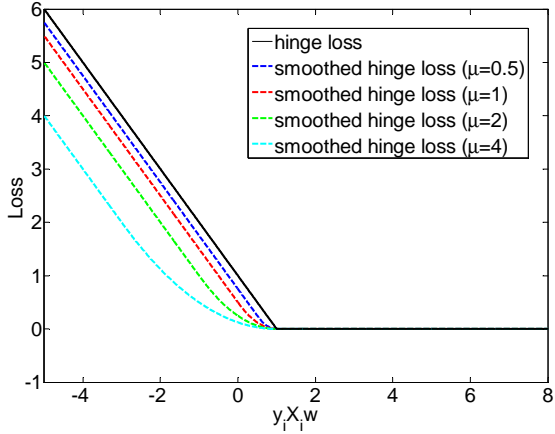}
\end{center}
   \caption{Hinge loss and smoothed hinge loss}
\label{fig:hingeloss}
\end{figure}

In SVM and LP-SVM, the loss function is given by the sum of all the hinge losses, i.e., $L(y_iX_i, w)=\sum_{i=1}^n \hbar \left(y_iX_i,w\right)$, which can be equivalently replaced by the following saddle point function,
\begin{align}
\min_{w\in \mathbb R^p}&\sum_{i=1}^n \hbar \left(y_iX_i,w\right)=\min_{w\in \mathbb R^p}\max_{u\in \mathcal Q}\langle
e-YXw,u\rangle,\\
\notag &\mathcal Q=\left\{u:0\leq u_i\leq 1,u\in \mathbb R^n\right\},
\end{align}
where $e$ is a vector full of $1$ and $Y={\rm Diag}(y)$. According to \cite{Nesterov}, the above saddle point function can be smoothed by subtracting a prox-function $d_1(u)$. The $d_1(u)$ is a strongly convex function of $u$ with a convex parameter $\sigma_1>0$ and the corresponding prox-center $u_0=\arg\min_{u\in \mathcal Q}d_1\left(u\right)$. Let $A_i$ be the $i^{th}$ row of the matrix $A$. We adopt $d_1\left(u\right)=\sum\nolimits_{i=1}^n\|X_i\|_\infty u_i^2$ in NESVM and thus the smoothed hinge loss $\hbar_\mu$ can be written as,
\begin{align}
\hbar_\mu=\max_{u\in \mathcal Q}u_i\left(1-y_iX_iw\right)-\frac{\mu}{2}\|X_i\|_\infty u_i^2,
\label{equ:hmu}
\end{align}
where $\mu$ is the smooth parameter. Since $d_1(u)$ is strongly convex, $u_i$ can be obtained by setting the gradient of the objective function in Eq.\ref{equ:hmu} as zero and then projecting $u_i$ on $\mathcal Q$, i.e.,
\begin{equation}
u_i={\rm median}\left\{\frac{1-y_iX_iw}{\mu\|X_i\|_\infty},0,1\right\}.
\label{equ:u2}
\end{equation}
Therefore, the smoothed hinge loss $\hbar_\mu$ is a piece-wise approximation of $\hbar$ according to different choices of $u_i$ in Eq.\ref{equ:u2}, i.e.,
\begin{align}\label{equ:u3}
\hbar_\mu=\left\{
  \begin{array}{ll}
    0, & \hbox{$y_iX_iw>1$;} \\
    \left(1-y_iX_iw\right)-\frac{\mu}{2}\|X_i\|_\infty, & \hbox{$y_iX_iw<1-\mu$;} \\
    \frac{\left(1-y_iX_iw\right)^2}{2\mu \|X_i\|_\infty}, & \hbox{else.}
  \end{array}
\right.
\end{align}
Fig.\ref{fig:hingeloss} plots the hinge loss $\hbar$ and smoothed hinge loss $\hbar_\mu$ with different $\mu$. The figure indicates that a larger $\mu$ induces a more smooth $\hbar_\mu$ with larger approximation error. The following theorem shows the theoretical bound of the approximation error.
\begin{theorem}
The hinge loss $\hbar$ is bounded by its smooth approximation $\hbar_\mu$, and the approximation error is completely controlled by the smooth parameter $\mu$. For any $w$, we have
\begin{equation}
\hbar_\mu\le \hbar \le \hbar_\mu+\frac{\mu}{2}\|X_i\|_\infty.
\end{equation}
\end{theorem}

According to Eq.\ref{equ:u2} and Eq.\ref{equ:u3}, the gradient of $\hbar_\mu$ for the $i^{th}$ sample is calculated as:
\begin{align}\label{equ:dh}
\notag \frac{\partial\hbar_\mu}{\partial w}&=\left\{
  \begin{array}{ll}
    0, & \hbox{$u_i=0$;} \\
    -\left(y_iX_i\right)^T, & \hbox{$u_i=1$;} \\
    \frac{-\left(y_iX_i\right)^T\cdot\left(1-y_iX_iw\right)}{\mu\|X_i\|_\infty}, & \hbox{$u_i=\frac{1-y_iX_iw}{\mu\|X_i\|_\infty}$.}
  \end{array}
\right\}\\
&=-\left(y_iX_i\right)^Tu_i.
\end{align}

In NESVM, the gradient of $L(y_iX_i, w)$ is used to determine the descent direction. Thus, the gradient of the sum of the smoothed hinge losses is given by
\begin{align}
\notag\frac{\partial L(y_iX_i, w)}{\partial w}=\frac{\partial\sum_{i=1}^n \hbar_\mu \left(y_iX_i,w\right)}{\partial w}=-\left(YX\right)^Tu.
\end{align}

In NESVM, the Lipschitz constant of $L(y_iX_i, w)$ is used to determine the step size of each iteration.

\begin{definition}
Given function $f(x)$, for arbitrary $x^1$ and $x^2$, Lipschitz constant $L$ satisfies
\begin{equation}
\|\nabla f(x^{1})-\nabla f(x^{2})\|_{2}\leq L\|x^{1}-x^{2}\|_{2}.
\end{equation}
\end{definition}
Thus the Lipschitz constant of $\hbar_\mu$ can be calculated from
\begin{equation}
\max\frac{\left\|\frac{\partial\hbar_\mu}{\partial w^1}-\frac{\partial\hbar_\mu}{\partial w^2}\right\|_2}{\left\|w^1-w^2\right\|_2}\leq L_{\hbar_\mu}.
\end{equation}
According to Eq.\ref{equ:dh}, we have
\begin{equation}
\frac{\partial\hbar_\mu}{\partial w^1}-\frac{\partial\hbar_\mu}{\partial w^2}=\left\{
  \begin{array}{lrr}
    0, ~~~~~~~\hbox{$y_iX_iw>1~{\rm or}~<1-\mu$;} \\
    \frac{X_i^TX_i\left(w^1-w^2\right)}{\mu\|X_i\|_\infty}, ~~~~~~~~~~~~~~~\hbox{else.}
  \end{array}
\right.
\end{equation}
Thus,
\begin{align}
\max\frac{\left\|X_i^TX_i\left(w^1-w^2\right)\right\|_2}{\mu\|X_i\|_\infty\left\|w^1-w^2\right\|_2}
\leq\frac{\left\|X_i^TX_i\right\|_2}{\mu\|X_i\|_\infty}=L_{\hbar_\mu}.
\end{align}
Hence the Lipschitz constant of $L(y_iX_i, w)$ (denoted as $L_\mu$) is calculated as
\begin{equation}\label{equ:LipHinge}
\sum_iL_{\hbar_\mu}\leq n\max_iL_{\hbar_\mu}=\frac{n}{\mu}\max_i\frac{\left\|X_i^TX_i\right\|_2}{\|X_i\|_\infty}=L_\mu.
\end{equation}

\subsection{Smooth the $\ell_1$-norm}

\begin{figure}[t]
\begin{center}
 \includegraphics[width=0.85\linewidth]{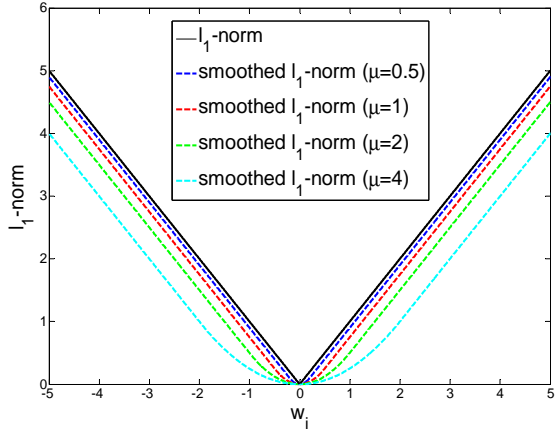}
\end{center}
   \caption{$\ell_1$-norm and smoothed $\ell_1$-norm}
\label{fig:l1norm}
\end{figure}

In LP-SVM, the regularizer is defined by the sum of all $\ell_1$-norm $\ell(w_i)=|w_i|$, i.e., $R(w)=\sum_{i=1}^p\ell(w_i)$. The minimization of $R(w)$ can be equivalently replaced by the following saddle point function,
\begin{align}
\min_{w\in \mathbb R^p}&\sum_{i=1}^p\ell(w_i)=\min_{w\in \mathbb R^p}\max_{u\in \mathcal Q}\langle w,u\rangle,\\
\notag &\mathcal Q=\left\{u:-1\leq u_i\leq1,u\in \mathbb R^p\right\}.
\end{align}
The above saddle point function can be smoothed by subtracting a prox-function $d_1(u)$. In this paper, we choose the prox-function $d_1\left(u\right)=(1/2)\|u\|_2^2$ and thus the smoothed $\ell_1$-norm $\ell_\mu$ can be written as,
\begin{align}
\ell_\mu\left(w_i\right)=\max_{u\in \mathcal Q}\langle w_i,u_i\rangle-\frac{\mu}{2}u_i^2.
\label{equ:lmu}
\end{align}
Since $d_1(u)$ is strongly convex, $u_i$ can be achieved by setting the gradient of the objective function in Eq.\ref{equ:lmu} as zero and then projecting $u_i$ on $\mathcal Q$, i.e.,
\begin{equation}
u_i={\rm median}\left\{\frac{w_i}{\mu},-1,1\right\},
\label{equ:u1}
\end{equation}
where $u$ can also be explained as the result of a soft thresholding of $w$. Therefore, the smoothed $\ell_1$-norm $\ell_\mu$ is a piece-wise approximation of $\ell$, i.e.,
\begin{align}
\notag \ell_\mu=\left\{
  \begin{array}{ll}
    -w_i-\frac{\mu}{2}, & \hbox{$w_i<-\mu$;} \\
    w_i-\frac{\mu}{2}, & \hbox{$w_i>\mu$;} \\
    \frac{w_i^2}{2\mu}, & \hbox{else.}
  \end{array}
\right.
\end{align}
Fig.\ref{fig:l1norm} plots the $\ell_1$-norm $\ell$ and the smoothed $\ell_1$-norm $\ell_\mu$ with different $\mu$. It shows that a larger $\mu$ induces a more smooth $\hbar_\mu$ with larger approximation error. The following theorem shows the theoretical bound of the approximation error.
\begin{theorem}
The $\ell_1$-norm $\ell$ is bounded by its smooth approximation $\ell_\mu$, and the approximation error is completely controlled by the smooth parameter $\mu$. For any $w$, we have
\begin{equation}
\ell_\mu\le \ell \le \ell_\mu+\frac{\mu}{2}.
\end{equation}
\end{theorem}

In NESVM, the gradient of $R(w)$ is used to determine the descent direction. Thus, the gradient of the sum of the smoothed $\ell_1$-norm $\ell_\mu$ is
\begin{align}
\frac{\partial\sum_{i=1}^p\ell_\mu(w_i)}{\partial w}=u.
\end{align}

In NESVM, the Lipschitz constant of $R(w)$ is used to determine the step size of each iteration. According to the definition of Lipschitz constant and the second order derivative of $\ell_\mu$ is given by
\begin{equation}
\frac{\partial^2\ell_\mu\left(w_i\right)}{\partial w_i^2}=\frac{1}{\mu},
\end{equation}
the Lipschitz constant of the sum of smoothed $\ell_1$-norm is given by
\begin{equation}\label{equ:LipLone}
L_\mu=\max_i\left\{\left|\frac{\partial^2\ell_\mu\left(w_i\right)}{\partial w_i^2}\right|\right\}=\frac{1}{\mu}.
\end{equation}

\subsection{Nesterov's method for SVM}

We apply Nesterov's method \cite{Nesterov} to minimize the smoothed primal SVM $F_\mu(w)$. It is a gradient method with the proved optimal convergence rate $\mathcal O(1/k^2)$. In its $k^{th}$ iteration round, two auxiliary optimizations are constructed and their solutions are used to build the SVM solution at the same iteration round. We use $w^k$, $y^k$ and $z^k$ to represent the solutions of SVM and its two auxiliary optimizations at the $k^{th}$ iteration round, respectively. The Lipschitz constant of $F_\mu(w)$ is $L_\mu$ and the two auxiliary optimizations are,
\begin{align}
\notag &\min_{y\in \mathbb{R}^{p}}\langle \nabla F_{\mu}(w^{k}),y-w^{k} \rangle+\frac{L_\mu}{2}\|y-w^{k}\|_{2}^{2},\\
\notag &\min_{z\in \mathbb{R}^{p}}\frac{L_\mu}{\sigma_2}d_2(z)+\sum_{i=0}^{k}\frac{i+1}{2}\left[F_\mu(w^i)+\langle \nabla F_{\mu}(w^{i}),z-w^{i}\rangle\right].
\end{align}
We choose the prox-function $d_2(z)=\|z-w^{\star}\|_{2}^{2}/2$ whose strong convexity parameter is $\sigma_2$, where $w^{\star}$ is the prox-center and $\sigma_2=1$. The $w^{\star}$ is usually selected as a guess solution of $w$.

By directly setting the gradients of the two objective functions in the auxiliary optimizations as zeros, we can obtain $y^k$ and $z^k$ respectively,
\begin{align}
&y^{k}=w^{k}-\frac{1}{L_\mu}\nabla F_{\mu}(w^{k}),
\label{equ:yk}
\end{align}\vspace{-5mm}
\begin{align}\vspace{-5mm}
&z^{k}=w^{\star}-\frac{\sigma_2}{L_\mu}\sum_{i=0}^{k}\frac{i+1}{2}\nabla F_{\mu}(w^{i}).
\label{equ:zk}
\end{align}
We have the following interpretation of the above results. The $y^k$ is a solution of the standard gradient descent with step size $1/L_\mu$ at the $k^{th}$ iteration round. The $z^k$ is a solution of a gradient descent step that starts from the guess solution $w^{\star}$ and proceeds along a direction determined by the weighted sum of negative gradients in all previous iteration rounds. The weights of gradients at later iteration rounds are larger than those at earlier iteration rounds. Therefore, $y^k$ and $z^k$ encode the current gradient and historical gradients. In NESVM, their weighted sum determines the SVM solution after the $k^{th}$ iteration round,
\begin{align}
w^{k+1}=\frac{2}{k+3}z^{k}+\frac{k+1}{k+3}y^{k}.
\label{equ:wk}
\end{align}
Let $\psi_k$ be the optimal objective value of the second auxiliary optimization, according to \cite{Nesterov}, we arrive at the following theorem.
\begin{theorem}
For any $k$ and the corresponding $y^k$, $z^k$ and $w^{k+1}$ defined by Eq.\ref{equ:yk}, Eq.\ref{equ:zk} and Eq.\ref{equ:wk}, respectively, we have
\begin{equation}
\frac{\left(k+1\right)\left(k+2\right)}{4}F_\mu\left(y^k\right)\leq \psi_k.
\end{equation}
\label{coretheorem}
\end{theorem}
Theorem \ref{coretheorem} is a direct result of Lemma 2 in \cite{Nesterov} and it will be applied to analyze the convergence rate of NESVM.

A small smooth parameter $\mu$ can improve the accuracy of the smooth approximation. A better guess solution $w^0$ that is close to the real one can improve the convergence rate and reduce the training time.

Algorithm 1 details the procedure of NESVM. In particular, the input parameters are the matrix $YX$, the initial solution $w^0$, the guess solution $w^\star$, the parameter $C$, the smooth parameter $\mu$ and the tolerance of termination criterion $\epsilon$. In each iteration round, the dual variable $u$ in smooth parts is first computed, then the gradient $\nabla F_\mu(w)$ is calculated from $u$, $y^k$ and $z^k$ are calculated from the gradient, and finally $w^{k+1}$ is updated at the end of the iteration round. NESVM conducts the above procedure iteratively until the convergence of $F_\mu\left(w\right)$.

\begin{algorithm}[!t]
\begin{algorithmic}
\STATE \textbf{Input:} $YX$, $w^0$, $w^{\star}$, $C$, $\mu$ and $\epsilon$
\STATE \textbf{Output:} weight vector $w$
\STATE \textbf{Initialize:} $k=0$
\REPEAT
\STATE Step 1: Compute dual variable $u$
\STATE Step 2: Compute gradient $\nabla F_\mu(w^k)$
\STATE Step 3: Compute $y^k$ and $z^k$ using Eq.\ref{equ:yk} and Eq.\ref{equ:zk}
\STATE Step 4: Update SVM solution $w^{k+1}$ using Eq.\ref{equ:wk}
\STATE Step 5: $k=k+1$
\UNTIL $|F_\mu(w^{k+1}) - F_\mu(w^k)| < \epsilon$
\STATE \textbf{return} $w=w^{k+1}$.
\end{algorithmic}
\caption{NESVM}
\end{algorithm}

NESVM contains no expensive computations, e.g., line search and Hessian matrix calculation. The most computational costs are two matrix-vector multiplications in Steps 1 and 2, i.e., $\left(YX\right)w$ and $\left(YX\right)^Tu$. Since most elements of $u$ are $0$ or $1$ and the proportion of these elements will rapidly increase with the decreasing of $\mu$, the computation of $\left(YX\right)^Tu$ can be further simplified. In addition, this simplification indicates that the gradient of each iteration round is completely determined by support vectors in the current iteration round. These support vectors correspond to the nonzero elements in $u$.

The above algorithm can be conveniently extended to nonlinear kernels by replacing the data matrix $X$ with $K\left(X,X\right)Y$, where $K\left(X,X\right)$ is the kernel matrix, and replacing the penalty $\|w\|_2^2$ with $w^TK(X,X)w$.

If a bias $b$ in an SVM classifier is required, let
\begin{align}
w:=\left[w;b\right]~~{\rm and}~~X:=\left[X,e\right],
\end{align}
Since the $\ell_2$ norm of $b$ is not penalized in the original SVM problem, we calculate the gradient of $b$ according to $\partial F(w)/\partial b=\partial L(y_iX_i, w)/\partial b$ in Algorithm 1, and the last entry of the output solution $w$ is the bias $b$.

\subsection{Convergence Analysis}

The following theorem shows the convergence rate, the iteration number and the time complexity of NESVM.
\begin{theorem}
The convergence rate of NESVM is $\mathcal O(1/k^2)$. It requires $\mathcal O(1/\sqrt{\epsilon})$ iteration rounds to reach an $\epsilon$ accurate solution.
\label{covergencetheorem}
\end{theorem}
\begin{proof}
Let the optimal solution be $w^*$. Since $F_\mu(w)$ is a convex function, we have
\begin{equation}
F_\mu(w^*)\geq F_\mu(w^i)+\langle\nabla F_\mu(w^i),w^*-w^i\rangle.
\end{equation}
Thus,
\begin{align}
\notag \psi_k&\leq\frac{L_\mu}{\sigma_2}d_2(w^*)+\sum_{i=0}^{k}\frac{i+1}{2}\left[F_\mu(w^i)+\langle \nabla F_{\mu}(w^{i}),w^*-w^{i}\rangle\right]\\
&\leq\frac{L_\mu}{\sigma_2}d_2(w^*)+\sum_{i=0}^k\frac{i+1}{2}F_\mu(w^*)\\
&=\frac{L_\mu}{\sigma_2}d_2(w^*)+\frac{\left(k+1\right)\left(k+2\right)}{4}F_\mu(w^*).
\end{align}
According to Theorem \ref{coretheorem}, we have
\begin{align}
&\frac{\left(k+1\right)\left(k+2\right)}{4}F_\mu\left(y^k\right)\leq\psi_k\leq\\
&\frac{L_\mu}{\sigma_2}d_2(w^*)+\frac{\left(k+1\right)\left(k+2\right)}{4}F_\mu(w^*).
\end{align}
Hence the accuracy at the $k^{th}$ iteration round is
\begin{equation}\label{equ:convergence}
F_\mu\left(y^k\right)-F_\mu(w^*)\leq \frac{4L_\mu d_2(w^*)}{\left(k+1\right)\left(k+2\right)}.
\end{equation}
Therefore, NESVM converges at rate $\mathcal O(1/k^2)$, and the minimum iteration number to reach an $\epsilon$ accurate solution is $\mathcal O(1/\sqrt{\epsilon})$. This completes the proof.
\end{proof}

According to the analysis in Section 2.3, there are only two matrix-vector multiplications in each iteration round of NESVM. Thus, the time complexity of each iteration round is $\mathcal O(n)$. According to Theorem \ref{covergencetheorem}, we can conclude the time complexity of NESVM is $\mathcal O(n/k^2)$.

\subsection{Accelerating NESVM with continuation}

The homotopy technique used in lasso \cite{lasso} and LARS \cite{LARS} shows the advantages of continuation method in speeding up the optimization and solving large-scale problems. In continuation method, a sequence of optimization problems with deceasing parameter is solved until the preferred value of the parameter is arrived. The solution of each optimization is used as the ``warm start'' for the next optimization. It has been proved that the convergence rate of each optimization is significantly accelerated by this technique, because only a few steps are required to reach the solution if the optimization starts from the ``warm start''.

In NESVM, a smaller smooth parameter $\mu$ is always preferred because it produces more accurate approximation of hinge loss or the $\ell_1$ norm. However, a small $\mu$ implies a large $L_\mu$ according to Eq.\ref{equ:LipHinge} and Eq.\ref{equ:LipLone}, which induces a slow convergence rate according to Eq.\ref{equ:convergence}. Hence the time cost of NESVM is expensive when small $\mu$ is selected.

We apply the continuation method to NESVM and obtain an accelerated algorithm termed ``homotopy NESVM'' for small $\mu$ situation. In homotopy NESVM, a series of smoothed SVM problems with decreasing smooth parameter $\mu$ are solved by using NESVM, and the solution of each NESVM is used as the initial solution $w^0$ of the next NESVM. The algorithm stops when the preferred $\mu=\mu^*$ is arrived. In this paper, homotopy NESVM starts from a large $\mu^0$, and sets the smooth parameter $\mu$ at the $t^{th}$ NESVM as
\begin{align}
\mu^t=\frac{\mu^0}{t+1}.
\label{equ:mu}
\end{align}
Because the smooth parameter $\mu$ used in each NESVM is large and the ``warm start'' is close to the solution, the computation of each NESVM's solution is cheap. In practice, less accuracy is often allowed for each NESVM, thus more computation can be saved. We show homotopy NESVM in Algorithm 2. Notice the Lipschitz constants in Eq.\ref{equ:LipHinge} and Eq.\ref{equ:LipLone} must be updated as the updating of the smooth parameter $\mu$ in Step 2.

\begin{algorithm}[!t]
\begin{algorithmic}
\STATE \textbf{Input:} $YX$, $w^0$, $w^{\star}$, $C$, $\mu^0$, $\epsilon$ and $\mu^*$.
\STATE \textbf{Output:} weight vector $w$.
\STATE \textbf{Initialize:} $t=0$.
\REPEAT
\STATE Step 1: Apply NESVM with $\mu=\mu^t$ and $w^0=w^t$
\STATE Step 2: Update $\mu^t=\mu^0/(t+1)$, $L_\mu$ and $t:=t+1$
\UNTIL $\mu^t\leq\mu^*$
\STATE \textbf{return} $w=w^t$.
\end{algorithmic}
\caption{Homotopy NESVM}
\end{algorithm}


\section{Examples}

In this section, we apply NESVM to three typical SVM models, i.e., classical SVM (C-SVM) \cite{CSVM}, linear programming SVM (LP-SVM) \cite{LPSVM} and least square (LS-SVM) \cite{LSSVM}\cite{SVMvsLSSVM}. They share an unified form Eq.\ref{equ:SVM}, and have different $R(w)$ and $L(y_iX_i, w)$. In NESVM, the solutions of C-SVM, LP-SVM and LS-SVM are different in calculating the gradient item $\nabla F_\mu(w^k)$ and the Lipschitz constant $L_\mu$.

\subsection{C-SVM}

In C-SVM, the regularizer $R(w)$ in Eq.\ref{equ:SVM} is
\begin{align}
R(w)=\frac{1}{2}\|w\|_2^2
\end{align}
and the loss function $L(y_iX_i, w)$ is the sum of all the hinge losses
\begin{align}
L(y_iX_i, w)=\sum_{i=1}^n \hbar \left(y_iX_i,w\right).
\end{align}
Therefore, the gradient $\nabla F_\mu(w^k)$ and the Lipschitz constant $L_\mu$ in NESVM are
\begin{align}
&\nabla F_\mu(w^k)=w^k-C\left(YX\right)^Tu,\\
&L_\mu=1+\frac{Cn}{\mu}\max_i\frac{\left\|X_i^TX_i\right\|_2}{\|X_i\|_\infty},
\end{align}
where $u$ is the dual variable in the smoothed hinge loss and can be calculated according to Eq.\ref{equ:u2}. Thus, C-SVM can be solved by using Algorithm 1 and Algorithm 2.

\subsection{LP-SVM}

In LP-SVM, the regularizer $R(w)$ in Eq.\ref{equ:SVM} is
\begin{align}
R(w)=\|w\|_1
\end{align}
and the loss function $L(y_iX_i, w)$ is the sum of all the hinge losses
\begin{align}
L(y_iX_i, w)=\sum_{i=1}^n \hbar \left(y_iX_i,w\right).
\end{align}
Therefore, the gradient $\nabla F_\mu(w^k)$ and the Lipschitz constant $L_\mu$ in NESVM are
\begin{align}
&\nabla F_\mu(w^k)=u-C\left(YX\right)^Tv,\\
&L_\mu=\frac{1}{\mu}+\frac{Cn}{\nu}\max_i\frac{\left\|X_i^TX_i\right\|_2}{\|X_i\|_\infty}.
\end{align}
where $u$ and $v$ are the dual variables in the smoothed $\ell_1$-norm and the smoothed hinge loss, and they can be calculated according to Eq.\ref{equ:u1} and Eq.\ref{equ:u2}, respectively. $\mu$ and $\nu$ are the corresponding smooth parameters. They are both updated according to Eq.\ref{equ:mu} with different initial values in homotopy NESVM. Thus LP-SVM can be solved by using Algorithm 1 and Algorithm 2.

\subsection{LS-SVM}

In LS-SVM, the regularizer $R(w)$ in Eq.\ref{equ:SVM} is
\begin{align}
R(w)=\frac{1}{2}\|w\|_2^2
\end{align}
and the loss function $L(y_iX_i, w)$ is the sum of all the quadratic hinge losses
\begin{align}
L(y_iX_i, w)=\sum_{i=1}^n \left(1-y_iX_iw\right)^2.
\end{align}
Since both the regularizer and the loss function are smooth, the gradient item $\nabla F(w^k)$ and the Lipschitz constant $L$ in NESVM are directly given by
\begin{align}
&\nabla F(w^k)=w^k-2C\left(YX\right)^T\left(1-YXw^k\right),\\
&L=1+2C\max_i\left\{\left\|X_i\right\|_2^2\right\}.
\end{align}
Steps 1 in Algorithm 1 is not necessary for LS-SVM, and thus LS-SVM can be solved by using Algorithm 1 and Algorithm 2.


\section{Experiments}

In the following experiments, we demonstrate the efficiency and effectiveness of the proposed NESVM by applying it to census income categorization and several computer vision tasks, i.e., indoor/outdoor scene classification, event recognition and scene recognition. We implemented NESVM in C++ and run all the experiments on a 3.0GHz Intel Xeon processor with 32GB of main memory under Windows Vista. We analyzed its scaling behavior and the sensitivity to $C$ and the size of dataset. Moreover, we compared NESVM against four benchmark SVM solvers, i.e., SVM-Perf \footnote{http://svmlight.joachims.org/svm\_perf.html}, Pegasos \footnote{http://www.cs.huji.ac.il/$\sim$ shais/code/index.html}, SVM-Light \footnote{http://svmlight.joachims.org} and LIBSVM \footnote{http://www.csie.ntu.edu.tw/$\sim$cjlin/libsvm/}. The tolerance used in stopping criteria of all the algorithms is set to $10^{-3}$. For all experiments, different SVM solvers obtained similar classification accuracies and performed comparably to the results reported in respective publications. Their efficiencies are evaluated by the training time in CPU seconds. All the SVM solvers are tested on $7$ different $C$ values, i.e., $\left\{10^{-3},10^{-2},10^{-1},1,10^1,10^2,10^3\right\}$ for $10$ times. We show their mean training time in following analysis. In the fist experiment, we also test the SVM solvers on $6$ subsets with different sizes.

Five experiments are exhibited, i.e., census income categorization, indoor scene classification, outdoor classification, event recognition and scene recognition. Linear C-SVM models are adopted in the first experiment to train binary classifies. Nonlinear C-SVM models with the RBF kernel $-\|X_i-X_j\|_2^2/p$ are adopted in the rest experiments, wherein $p$ is the number of features. For multiclass classification tasks, the one-versus-one method was adopted. Pegasos is compared with NESVM in the first experiment, because its code is only available to linear C-SVM. In all the experiments, we set the initial solution $w^0=\textbf{0}$, the guess solution $w^\star=\textbf{0}$, the smooth parameter $\mu=5$ and the tolerance of termination criterion $\epsilon=10^{-3}$ in NESVM.

\subsection{Census income categorization}

We consider the census income categorization on the Adult dataset from UCI machine learning repository \cite{UCI}. The Adult contains $123$ dimensional census data of $48842$ Americans. The samples are separated into two classes according to whether their income exceeds $\$50K/yr$ or not. Table 1 shows the number of training samples and the number of test samples in each subset.

\begin{table}
\begin{center}
\begin{tabular}{|l||@{}c@{}|@{}c@{}|@{}c@{}|@{}c@{}|@{}c@{}|@{}c@{}|}
\hline
Set ID &1 &2 &3 &4 &5 &6 \\
\hline
Training set &~1605~ &~2265~ &~3185~ &~4781~ &~6414~ &~11220~ \\
\hline
Test set &~30956~ &~30296~ &~29376~ &~27780~ &~26147~ &~21341~ \\
\hline
\end{tabular}
\end{center}
\caption{Six subsets in the census income dataset.}
\end{table}

\begin{figure}[t]
\begin{center}
 \includegraphics[width=0.9\linewidth]{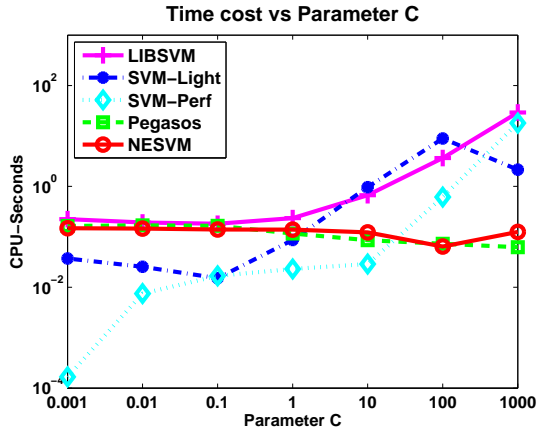}
\end{center}
   \caption{Time cost vs $C$ in census income categorization}
\label{fig:adult1}
\end{figure}

Fig.\ref{fig:adult1} shows the scalability of the five SVM solvers on the different $C$ values. The training time of NESVM and Pegasos is slightly slower than the other SVM solvers for small $C$ and faster than the others for large $C$. In addition, NESVM and Pegasos are least sensitive to $C$, because the search of the most violated constraint in SVM-Perf, and the working set selection in SVM-Light and LIBSVM will be evidently slowed when $C$ is augmented. However, the main computations of NESVM and Pegasos are irrelevant to $C$.

\begin{figure}[t]
\begin{center}
 \includegraphics[width=0.9\linewidth]{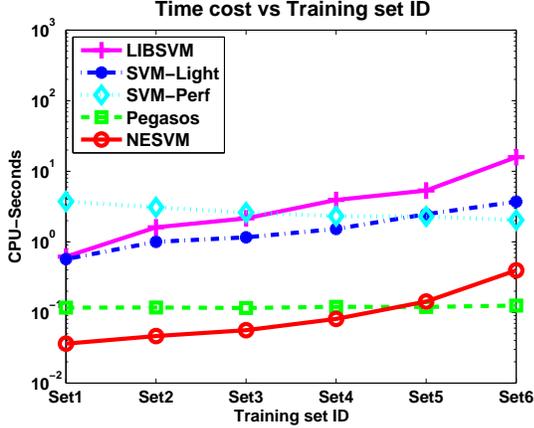}
\end{center}
   \caption{Time cost vs set ID in census income categorization}
\label{fig:adult2}
\end{figure}

Fig.\ref{fig:adult2} shows the scalability of the five SVM solvers on subsets with increasing sizes. NESVM achieves the shortest training time when the number of training samples is less than $5000$. Moreover, NESVM and Pegasos are least sensitive to the data size among all the SVM solvers. Pegasos achieves shorter training time when the number of training samples is more than $10000$, this is because NESVM is a batch method while Pegasos is an online learning method.

\subsection{Indoor scene classification}

\begin{figure*}
\begin{center}
 \includegraphics[width=0.85\linewidth]{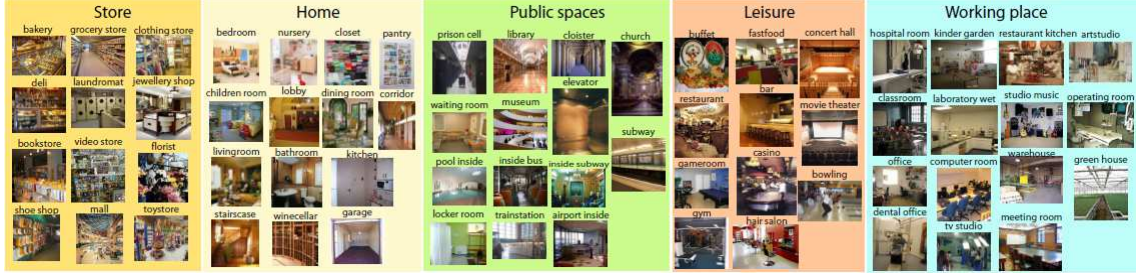}
\end{center}
   \caption{Sample images of indoor scene dataset}
\label{fig:sindoor}
\end{figure*}

We apply NESVM to indoor scene classification on the dataset proposed in \cite{Indoor_scene}. The minimum resolution of all images in the smallest axis are $200$ pixels. The sample images are shown in Fig.\ref{fig:sindoor}. We choose a subset of the dataset by randomly selecting $1000$ images from each of the five given groups, i.e., store, home, public spaces, leisure and working place. Gist features of $544$ dimensions composed of color, texture and intensity are extracted to represent images. In our experiment, $70\%$ data are randomly selected for training, and the rest for testing.

\begin{figure}[t]
\begin{center}
 \includegraphics[width=0.9\linewidth]{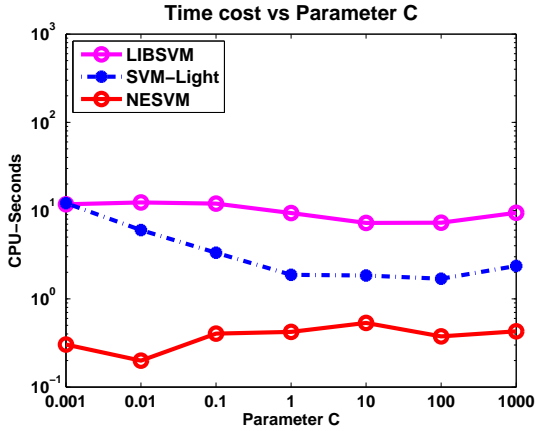}
\end{center}
   \caption{Time cost vs $C$ in indoor scene classification}
\label{fig:indoor}
\end{figure}

Fig.\ref{fig:indoor} shows the scalability of four SVM solvers on the different $C$ values. NESVM achieves the shortest training time on different $C$ among all the SVM solvers, because NESVM obtains the optimal convergence rate $\mathcal O(1/k^2)$ in its gradient descent. LIBSVM has the most expensive time cost among all the SVM solvers. In addition, NESVM is least sensitive to $C$, because the main calculations of NESVM, i.e., the two matrix-vector multiplications, are irrelevant to $C$. SVM-Light is most sensitive to $C$. SVM-Perf is not shown in Fig.\ref{fig:indoor} because its training time is much more than the other SVM solvers on all the $C$ (more than 1000 CPU seconds).

\subsection{Outdoor scene classification}

\begin{figure}[t]
\begin{center}
 \includegraphics[width=0.9\linewidth]{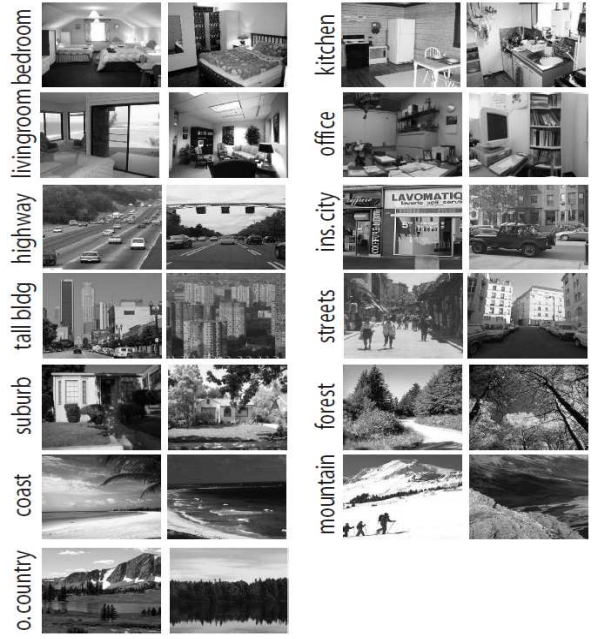}
\end{center}
   \caption{Sample images of outdoor scene dataset}
\label{fig:soutdoor}
\end{figure}

We apply NESVM to outdoor scene classification on the dataset proposed in \cite{Outdoor_scene}. It contains $13$ classes of natural scenes, e.g., highway, inside of cities and office. The sample images are shown in Fig.\ref{fig:soutdoor}. Each class includes $200$-$400$ images, we split the images into $70\%$ training samples and $30\%$ test samples. The average image size is $250\times 300$ pixels. Gist features of $352$ dimensions composed of texture and intensity are extracted to represent grayscale images.

\begin{figure}[t]
\begin{center}
 \includegraphics[width=0.9\linewidth]{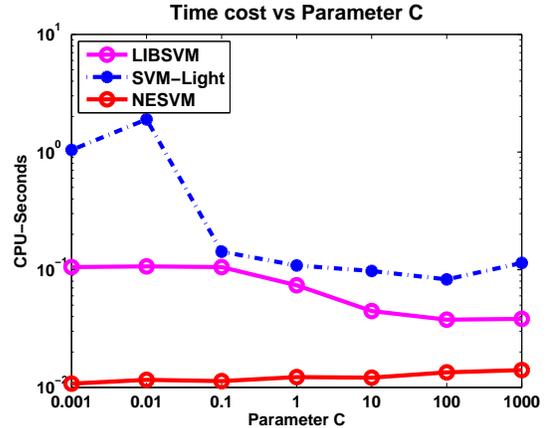}
\end{center}
   \caption{Time cost vs $C$ in outdoor scene classification}
\label{fig:outdoor}
\end{figure}

Fig.\ref{fig:outdoor} shows the scalability of the four SVM solvers on the different $C$ values. NESVM is more efficient than SVM-Light and LIBSVM. It took more than 100 CPU seconds for SVM-Perf on each $C$, so we do not show SVM-Perf.

\subsection{Event recognition}

\begin{figure}[t]
\begin{center}
 \includegraphics[width=0.9\linewidth]{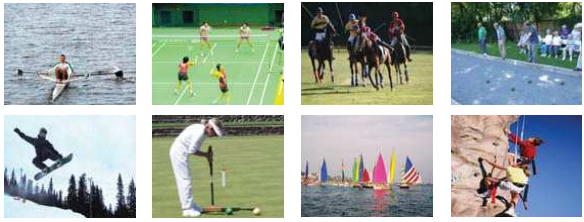}
\end{center}
   \caption{Sample images of event dataset}
\label{fig:sevent}
\end{figure}

We apply NESVM to event recognition on the dataset proposed in \cite{Events}. It contains $8$ classes of sports events, e.g., bocce, croquet and rock climbing. The size of each class varies from $137$ to $250$. The sample images are shown in Fig.\ref{fig:sevent}. Bag of words features of $300$ dimensions are extracted according to \cite{Events}. We split the dataset into $70\%$ training samples and $30\%$ test samples.

\begin{figure}[t]
\begin{center}
 \includegraphics[width=0.9\linewidth]{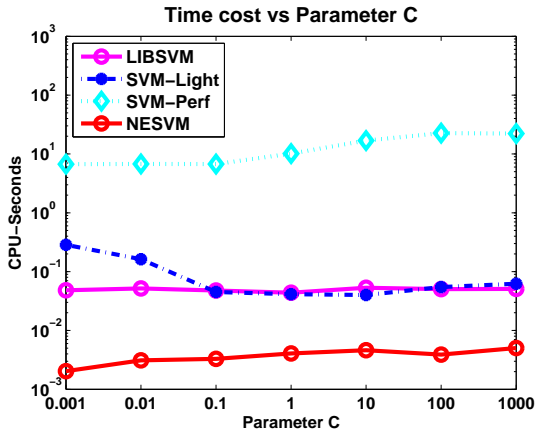}
\end{center}
   \caption{Time cost vs $C$ in event recognition}
\label{fig:event}
\end{figure}

Fig.\ref{fig:event} shows the scalability of the four SVM solvers on the different $C$ values. NESVM achieves the shortest training time on different $C$ among all the SVM solvers. SVM-Light and LIBSVM have similar CPU seconds, because both of them are based on SMO. SVM-Perf has the most expensive time cost on different $C$, because advantages of the cutting-plane algorithm used in SVM-Perf are weakened in the nonlinear kernel situation. NESVM and LIBSVM are less sensitive to $C$ than SVM-Perf and SVM-Light.

\subsection{Scene recognition}

We apply NESVM to scene recognition on the dataset proposed in \cite{Nus_Wide}. It contains $6$ classes of images, i.e., event, program, scene, people, objects and graphics. We randomly select $10000$ samples from the scene class and $10000$ samples from the other classes and obtain a dataset with $20000$ samples. Bag of words features of $500$ dimensions are extracted according to \cite{Nus_Wide}. We split the dataset into $50\%$ training samples and $50\%$ test samples.

\begin{figure}[t]
\begin{center}
 \includegraphics[width=0.9\linewidth]{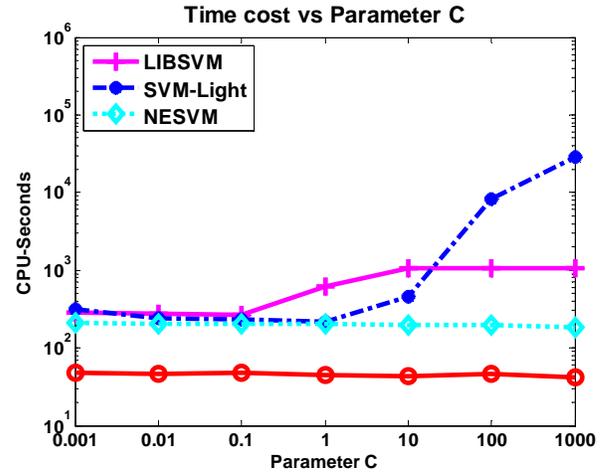}
\end{center}
   \caption{Time cost vs $C$ in scene recognition}
\label{fig:scene}
\end{figure}

Fig.\ref{fig:scene} shows the scalability of the four SVM solvers on the different $C$ values. NESVM achieves the shortest training time on different $C$ among all the SVM solvers. The training time of SVM-Light and LIBSVM similarly increase as the augment of $C$, because both of them are based on SMO. NESVM and SVM-Perf are less sensitive to $C$ than LIBSVM and SVM-Light in this binary classification.

\section{Conclusion}

This paper presented NESVM to solve the primal SVMs, e.g., classical SVM, linear programming SVM and least square SVM, with the optimal convergence rate $\mathcal O(1/k^{2})$ and a linear time complexity. Both linear and nonlinear kernels can be easily applied to NESVM. In each iteration round of NESVM, two auxiliary optimizations are constructed and a weighted sum of their solutions are adopted as the current SVM solution, in which the current gradient and the historical gradients are combined to determine the descent direction. The step size is automatically determined by the Lipschitz constant of the objective. Two matrix-vector multiplications are required in each iteration round.

We propose an accelerated NESVM, i.e., homotopy NESVM, to improve the efficiency of NESVM when accurate approximation of hinge loss or the $\ell_1$ norm is required. Homotopy NESVM solves a series of NESVM with decreasing smooth parameter $\mu$, and the solution of each NESVM is adopted as the ``warm start'' of the next NESVM. The time cost caused by small $\mu$ and the starting point $w^0$ far from the solution can be significantly saved by using homotopy NESVM.

The experiments on various applications indicate that NESVM achieves the competitive efficiency compared against four popular SVM solvers, i.e., SVM-Perf, Pegasos, SVM-Light and LIBSVM, and it is insensitive to $C$ and the size of dataset. NESVM can be further studied in many areas. For example, it can be sophisticatedly refined to handle sparse features in document classification. Its efficiency can be further improved by introducing the parallel computation. Because the gradient of the smoothed hinge loss and the smoothed $\ell_1$ norm is already obtained, NESVM can be further accelerated by extending it to online learning or stochastic gradient algorithms. These will be mainly studied in our future work.

{\small
\bibliographystyle{IEEEtran}
\bibliography{NESVM_ref}
}

\end{document}